\documentclass[10pt,twocolumn,letterpaper]{article}

\usepackage[accsupp]{axessibility}  \usepackage{cvpr}

\definecolor{cvprblue}{rgb}{0.21,0.49,0.74}
\usepackage[pagebackref,breaklinks,colorlinks,allcolors=cvprblue, linkcolor=red]{hyperref}

\usepackage{booktabs}
\usepackage[table]{xcolor}
\usepackage{siunitx}
\usepackage{multirow}
\usepackage[T1]{fontenc}
\usepackage{algorithm}       \usepackage{algpseudocode}   

\usepackage{xcolor}
\definecolor{darkgreen}{RGB}{0,100,0} 

\def\paperID{} \def\confName{CVPR}
\def\confYear{2026}

\title{Back to the Feature: Explaining Video Classifiers with Video Counterfactual Explanations}

\author{
Chao Wang\textsuperscript{1,2}\quad
Chengan Che\textsuperscript{1,2}\quad
Xinyue Chen\textsuperscript{1,2} \quad
Sophia Tsoka\textsuperscript{2}\quad
Luis C. Garcia-Peraza-Herrera\textsuperscript{1,2} \\
\vspace{2pt} \\
\textsuperscript{1}Visual Understanding Research Group, Department of Informatics, King’s College London, UK \\
\textsuperscript{2}Department of Informatics, King’s College London, UK
}
\usepackage{placeins}
\begin{document}
\maketitle
\begin{abstract}
Counterfactual explanations (CFEs) are minimal and semantically meaningful modifications of the input of a model that alter the model predictions.
They highlight the decisive features the model relies on, providing contrastive interpretations for classifiers. 
State-of-the-art visual counterfactual explanation methods have primarily focused on interpreting image classifiers, leaving the domain of video models relatively underexplored.
For the video CFEs to be useful, they have to be physically plausible, temporally coherent, and exhibit smooth motion trajectories. 
Existing CFE image-based methods, designed to explain image classifiers, lack the capacity to generate temporally coherent, smooth and physically plausible video CFEs. To address this, we propose \textbf{Back To The Feature (BTTF)}, an optimization framework that generates video CFEs.
Our method introduces two novel features, 1) an optimization scheme to retrieve the initial latent noise conditioned by the first frame of the input video, 2) a two-stage optimization strategy to enable the search for counterfactual videos in the vicinity of the input video.
Both optimization processes are guided solely by the target classifier, ensuring the explanation is faithful.
To accelerate convergence, we also introduce a progressive optimization strategy that incrementally increases the number of denoising steps.
Extensive experiments on video datasets such as Shape-Moving (motion classification), MEAD (emotion classification), and NTU RGB+D (action classification) show that our BTTF effectively generates valid, visually similar and realistic counterfactual videos that provide concrete insights into the classifier's decision-making mechanism. 
See our project page for code and more results: \href{bttf.visurg.ai}{https://bttf.visurg.ai}.
\end{abstract}

 \section{Introduction}
\label{sec:intro}

\begin{figure}[t]
  \centering
  \includegraphics[width=\linewidth]{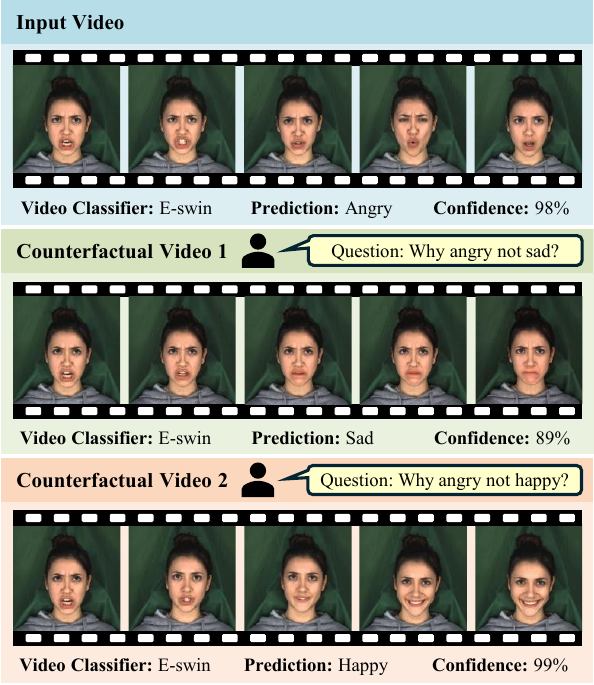}
   \caption{\textbf{Video counterfactual explanations with BTTF.}
    The top row shows an input video, where the facial expression is predicted as ``Angry'' with 98\% confidence by the target video classifier E-swin. 
    To answer ``why angry not sad?'', our method BTTF (middle row) introduces minimal and semantically meaningful changes to the input video, resulting in the alteration of the model's prediction to ``Sad''. Similarly, to answer ``why angry not happy?'', another counterfactual video classified as ``Happy'' is generated (bottom row). These counterfactual explanations visually highlight the key spatiotemporal features (facial movements) that the classifier relies on to make its decisions.
   }
   \label{fig:teaser}
\end{figure}

Modern computer vision models, while powerful, are increasingly opaque, making their internal logic difficult to interpret.
This lack of transparency poses a significant risk in critical domains, where predictive errors due to hidden biases may lead to disastrous outcomes~\cite{eykholt2018robust,weng2024fast}.
Counterfactual Explanations (CFEs)~\cite{wachter2017counterfactual, guidotti2024counterfactual} address this challenge by asking a \textit{what if} type of question: ``What minimal and semantically-meaningful changes to an original input are needed for the model to alter its decision from the original prediction to a new, target prediction?''
By identifying these adjustments, CFEs offer insight into why the model originally produced a different result.

While much work has demonstrated the strong performance of CFEs in explaining the decision-making mechanism of image classifiers~\cite{boreiko2022sparse, jeanneret2022diffusion, augustin2022diffusion, weng2024fast, augustin2024dig, jeanneret2024text, sobieski2024rethinking, farid2023latent}, particularly detecting failure modes (e.g., spurious features and shortcuts)~\cite{boreiko2022sparse, weng2024fast},
extending the concept of CFEs to the field of video classification remains an underexplored and challenging task.
Existing image-based methods primarily perturb static features (e.g., textures or colors), which is insufficient for videos~\cite{boreiko2022sparse}.
Unlike static images, most video features (e.g., object motion, facial emotion and human action) are dynamic and have both spatial and temporal attributes~\cite{laptev2005space, ji20123d}. 
Consequently, editing videos not only requires changing both local and global features inside a single frame, but also necessitates coordinated, temporally consistent adjustments across all frames to preserve physical plausibility and narrative continuity~\cite{chu2020learning, wang2018video}.
These restrictions impose additional demands on CFE tasks. 

Based on definitions of CFEs in the literature~\cite{Verma2020CounterfactualEF, boreiko2022sparse}, we summarize video CFEs as a video editing task that should satisfy five criteria: 
\textbf{(i) validity}, the \textit{target model} predicts the \textit{specified target label} on the \textit{generated counterfactual video};
\textbf{(ii) proximity}, \textit{the edit} is minimal, localized and confined to designated semantic factors (e.g., actions) while preserving scene identity and layout;
\textbf{(iii) actionability}, \textit{the edit} belongs to \textit{a feasible set} of manipulations that human users could plausibly enact, ruling out unattainable pixel-level tweaks; 
\textbf{(iv) realism}, the \textit{generated counterfactual video} lies on the natural-video manifold consistent with the \textit{original input}, exhibiting photorealistic appearance and coherent dynamics; and 
\textbf{(v) spatiotemporal-consistency}, \textit{edits} must maintain coherent dependencies among spatial and temporal features, yielding physically plausible, temporally smooth trajectories without violating scene semantics.

In this work, we propose \textbf{Back To The Feature (BTTF)}, an optimization framework where Image-To-Video (I2V) diffusion models are employed as the generator to synthesize counterfactuals for explaining video classifiers. 
Specifically, given an \textit{original input video}, our proposed method searches for a \textit{new video} which starts at the same moment as the \textit{original}.
Drawing inspiration from the ``many-worlds'' interpretation of quantum mechanics \cite{everett1957},
the \textit{generated video} can be viewed as the closest spatiotemporally parallel video where a small number of changed dynamic features results in the alteration of the target classifier's predictions.
As shown in~\cref{fig:teaser}, comparing the difference between generated counterfactuals and original inputs gives us insights into the inner workings of video classifiers. 

Our method incorporates several key designs that directly address the five criteria that a CFE should satisfy:  
\textbf{(i) validity}, the classifier’s cross-entropy loss with the specified target label is used as the main driving signal during the optimization process, ensuring the resulting video is classified as intended; 
\textbf{(ii) proximity}, we adopt a two-stage scheme in which an inversion stage first anchors the search near the \textit{original video}, providing a close initial state for subsequent counterfactual optimization;  
\textbf{(iii) actionability}, optimization occurs in the semantically-rich VAE latent space, so all edits are mapped by the diffusion model to coherent video-level semantic attributes rather than meaningless noise; 
\textbf{(iv) realism}, conditioning the generator on the original first frame, together with a translation-invariant video style regularizer, constrains the optimization so that the generated counterfactual remains on the input video manifold; 
\textbf{(v) spatiotemporal-consistency}, diffusion models are pre-trained on large-scale video datasets and fine-tuned on video data from the same domain as the target classifier, enabling the synthesis of dynamics that respect physical and temporal regularities. 

Our contributions can be summarized as follows: 
\begin{itemize}
    \item We propose \textbf{BTTF}, a novel optimization framework capable of editing actionable spatiotemporal features to generate CFEs for video classifiers.
\item We design a two-stage optimization procedure that promotes the \textbf{proximity} of generated counterfactual videos to the original video.
\item We introduce a conditioning strategy using the original first frame and style loss regularization to produce \textbf{realistic} and \textbf{spatiotemporally-consistent} videos.
\item We propose a progressive optimization strategy to accelerate the convergence of the counterfactual generation optimization process.
\item Our counterfactual generation process is guided only by the target classifier, ensuring \textbf{faithfulness} to the classifier’s internal decision mechanism.
\item We demonstrate experimentally that BTTF provides useful insights into the decision-making logic of video classifiers and can effectively identify spurious features within them.
\end{itemize}

\begin{figure*}[!t]
  \centering
  \includegraphics[width=1.0\linewidth]{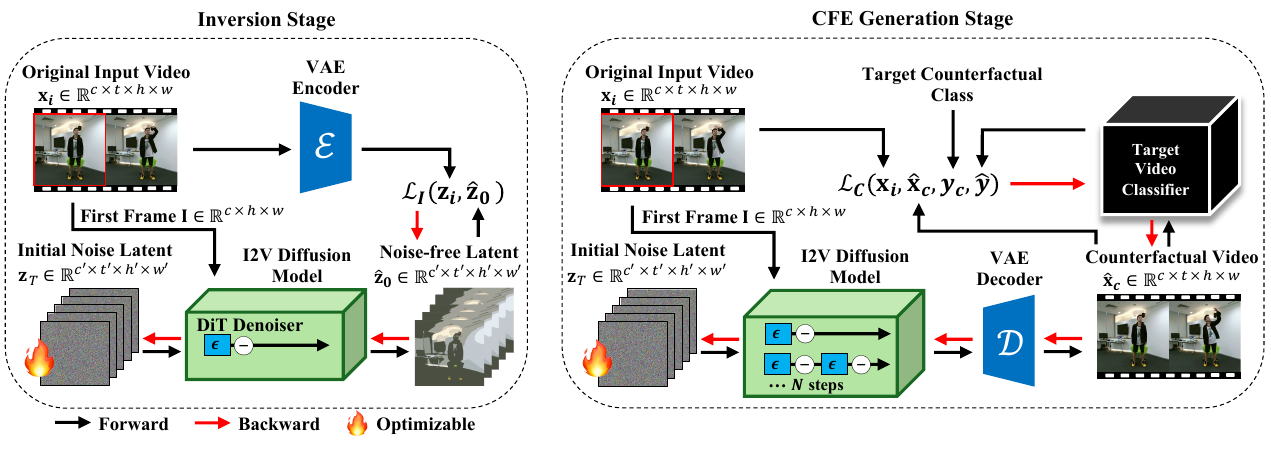}
  \caption{\textbf{Illustration of the BTTF optimization framework for video CFEs.} In the first stage for inversion, the initial latent input $\mathbf{z}_T$, which is initially sampled from the Gaussian distribution, is optimized by the backpropagated gradients from the reconstruction loss $\mathcal{L}_I$ between the noise-free latent $\hat{\mathbf{z}}_0$ and the original input video latent $\mathbf{z}_i$. In the second stage for CFE generation, $\hat{\mathbf{z}}_0$ is further decoded by the VAE decoder $\mathcal{D}$ to obtain the generated counterfactual video $\hat{\mathbf{x}}_c$, and then fed into the target video classifier (black box) to compute the cross-entropy loss with the target class $y_c$. The cross-entropy loss, together with the video style loss computed from $\hat{\mathbf{x}}_c$ and $\mathbf{x}_i$, constitutes the objective function $\mathcal{L}_C$, which is used to optimize $\mathbf{z}_T$ via gradient backpropagation. In the denoising process of our I2V diffusion models, the number of inference steps is set to one in the first stage, while it progressively increases from one to $N$ in the second stage to accelerate the convergence of the optimization process.}
  \label{fig:scheme}
\end{figure*}

 \section{Related Work}
\label{sec:related}

In this section, we present related work about CFEs.
Notably, counterfactual explanation and counterfactual generation are related but separate research areas.
Counterfactual explanations aim to elucidate a model’s decision-making process~\cite{guidotti2024counterfactual}.
In contrast, counterfactual generation focuses on creating alternative versions of data,
rather than explaining a model~\cite{komanduri2023identifiable}.
While there already exists work about video counterfactual generation~\cite{reynaud2022Dartagnan,venkatesh2024understanding}, video CFE is still an under-explored area.

\subsection{Image Counterfactual Explanations}

Existing image-based CFE approaches can be broadly categorized into two families: (i) adversarial attack-based and (ii) generative model-based methods.

Adversarial attacks~\cite{madry2017towards, chen2020frank} are inherently related to CFEs.
Both involve perturbing the input in order to change the model's prediction.
However, perturbations of adversarial attacks are usually noisy and non-actionable~\cite{jeanneret2023adversarial}.
In contrast, CFEs aim to reveal causal relationships between input features and model predictions by applying semantically meaningful modifications.
Despite this difference, prior work~\cite{santurkar2019image} shows that attacking adversarially robust classifiers would induce meaningful perturbations to the input image.
Therefore, pioneering work~\cite{boreiko2022sparse} of image-based CFEs directly adapted adversarial attack methods to generate CFEs for robust image classifiers.
However, no studies indicate that this can also work in the video setting. 

Many modern frameworks for producing CFEs employ generative models, such as variational autoencoders (VAEs)~\cite{rodriguez2021beyond}, generative adversarial networks (GANs)~\cite{jacob2022steex, khorram2022cycle, luo2023zero} and diffusion models~\cite{jeanneret2022diffusion,augustin2022diffusion,farid2023latent}. 
Due to stable training and high-quality image generation, diffusion models have become the preferred CFE generators in recent work~\cite{augustin2024dig,weng2024fast,sobieski2024rethinking}. 
The central problem is how to effectively fuse the input image and classifier information within the diffusion process to synthesize realistic CFEs. 
A widely adopted strategy~\cite{jeanneret2022diffusion,augustin2022diffusion,weng2024fast,sobieski2024rethinking,farid2023latent} is to partially corrupt the input image with noise to preserve the coarse structure of the input image. Then classifier guidance (CG)~\cite{dhariwal2021diffusion} is incorporated to steer the denoising trajectory toward the target class. 
Although intuitive, this strategy introduces two important limitations.
First, beginning the generation from an incompletely noised input constrains the diffusion model to operate primarily in low-noise stages of the trajectory, which correspond to stages dominated by local texture formation rather than global structural change~\cite{park2023understanding,liu2024structure}.
Consequently, such a design is ill-suited when the required counterfactual involves perturbing spatiotemporal features, such as object motion patterns or human actions in videos, because editing late denoising stages is insufficient to modify motion or large structural changes~\cite{wu2023tune,khachatryan2023text2video}.
Second, when guidance is derived from adversarially non-robust classifiers, diffusion models often produce noisy and non-meaningful perturbations (adversarial examples)~\cite{weng2024fast,nichol2021glide}.
Augustin et al.~\cite{augustin2022diffusion} attribute this problem to the inherently noisy and less informative gradients of non-robust classifiers. To address this, they propose DVCE~\cite{augustin2022diffusion}, which introduces a cone-projection mechanism to align the gradient of the non-robust classifier with that of an auxiliary robust classifier.
While this alignment encourages the diffusion process to yield meaningful perturbations, it also raises a fundamental concern: the counterfactual edits depend not only on the target classifier but also on the robust auxiliary model. As a result, the generated counterfactuals may no longer faithfully reflect the internal decision logic of the target classifier.
Distinct from classifier-guided diffusion approaches, UVCE~\cite{augustin2024dig} adopts an optimization-guided diffusion method. 
Specifically, it uses a Text-To-Image (T2I) diffusion model (Stable Diffusion)~\cite{rombach2022high} as the counterfactual generator and optimizes its inputs (namely the initial latent and the prompt embeddings) via the gradients. 
Although UVCE can synthesize realistic counterfactuals, it explicitly embeds the target class name in the initial prompt, meaning that the resulting CFEs are shaped not only by classifier feedback but also by prior class-specific knowledge learned during T2I training.

\subsection{Video Counterfactual Explanations} 

Research on CFEs for video models remains very limited. 
A closely related early work~\cite{kanehira2019multimodal} provides explanations by identifying attributes and spatiotemporal regions in a video that can push the classifier's prediction toward a counterfactual label.
While insightful, this method does not generate counterfactual videos, which distinguishes it from modern counterfactual explanations~\cite{wachter2017counterfactual}.
Overall, video CFE is still an underexplored area.
While image-based CFEs have achieved remarkable progress, their adaptation to videos remains technically challenging due to the spatiotemporal nature of videos.
In addition, existing image-based approaches~\cite{augustin2022diffusion,augustin2024dig,farid2023latent} often compromise the faithfulness of CFEs by relying on auxiliary guidance or text prompts. 
These limitations highlight the need for a principled video CFE framework that can (i) manipulate spatiotemporal features of input videos in an actionable manner and (ii) faithfully reflect the target classifier's internal reasoning mechanism.

 \section{Methodology}
\label{sec:meth}

\begin{figure}[t]
  \centering
  \includegraphics[width=\linewidth]{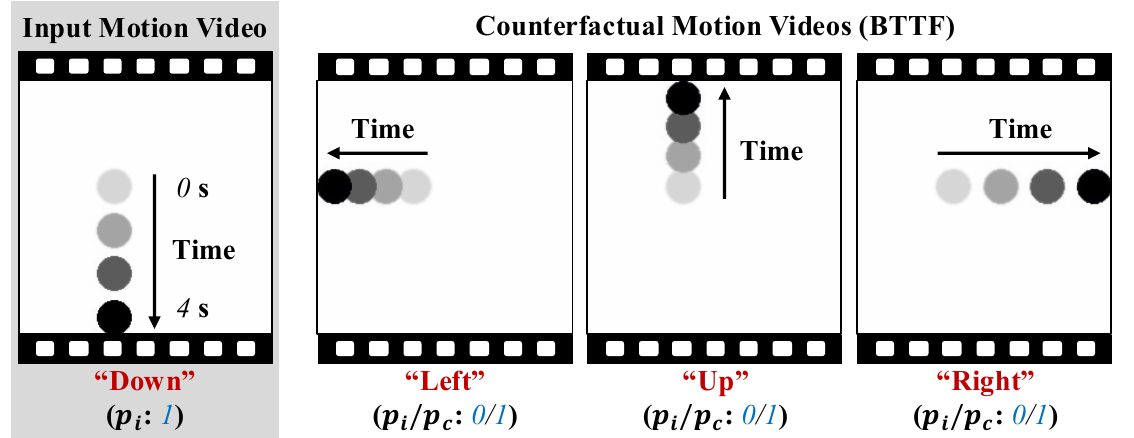}
   \caption{\textbf{CFE videos generated by BTTF for the target motion classifier M-swin trained on Shape-Moving.} 
BTTF changes M-swin's prediction on the original input video from ``Down'' to target motion classes ``Left'', ``Up''  and ``Right'', respectively.  $p_i$ denotes the confidence of the target classifier for the original class of input video, and $p_c$ denotes the confidence for the target class. This notation is consistently used in the subsequent figures.}
   \label{fig:shape_moving_CFE}
\end{figure}

\begin{table}
  \caption{\textbf{Performance evaluation of Video Swin Transformer-based target classifiers.} Acc. is the standard accuracy, while RA is the robust accuracy tested under $l_2$-PGD attack (perturbation budget $\epsilon$ = 20, number of perturbation steps $n$ = 10).}
  \label{tab:classifiers_eval}
  \centering
  \begin{tabular}{@{}llll@{}}
    \toprule
    Training set & Classifier & Acc. & RA \\
    \midrule
    \multirow{1}{*}{Shape moving}
      & M-swin    & 100.0                     & 28.8 \\
    \addlinespace
    \multirow{1}{*}{MEAD}
      & E-swin    & 98.9                    & 0 \\
\addlinespace
    \multirow{1}{*}{NTU RGB+D}
& A-swinR    & 54.2                    & 35.8 \\
\bottomrule
  \end{tabular}
\end{table}

\subsection{Preliminaries: I2V Diffusion Models}

Score-based diffusion models (SBDMs)~\cite{ho2020denoising, song2019generative, song2020improved} generate samples by progressively denoising an initial noise variable $\mathbf{x}_T$ drawn from a simple noise distribution $p(\mathbf{x}_T)$ (typically Gaussian).
The process produces a sequence of intermediate variables $(\mathbf{x}_t)_{t=T}^0$, eventually yielding a clean sample $\mathbf{x}_0$ (e.g., an image or a video).
Instead of operating directly in pixel space, latent diffusion models~\cite{rombach2022high} perform denoising in the latent space of a VAE, gradually refining a noisy latent $\mathbf{z}_T$ into a clean latent $\mathbf{z}_0$, which is then decoded into a final sample $\mathbf{x}_0 = \mathcal{D}(\mathbf{z}_0)$.
Furthermore, diffusion models can incorporate auxiliary conditioning (e.g., text or image)~\cite{wan2025wan, kong2024hunyuanvideo, agarwal2025cosmos} to control the denoising trajectory.

In this work, we adopt Wan-I2V series~\cite{wan2025wan}, a state-of-the-art I2V latent diffusion model. 
Given an input image $\mathbf{I}$, the model injects information along two parallel pathways: 
(i) CLIP~\cite{radford2021learning} image features are extracted and fused into a diffusion transformer (DiT)-based~\cite{peebles2023scalable} denoiser $\epsilon_{\boldsymbol{\phi}}$ via cross-attention, 
and 
(ii) the VAE-encoded image latent $\mathcal{E}(\mathbf{I})$ is concatenated with an initial Gaussian noise latent $\mathbf{z}_T$ and jointly processed by $\epsilon_{\boldsymbol{\phi}}$.
During inference, Wan-I2V employs a deterministic flow matching sampler~\cite{lipman2022flow,holderrieth2025introduction}.
For a fixed input image $\mathbf{I}$, the final output video $\mathbf{x}_0$ becomes a deterministic function of the initial noise latent $\mathbf{z}_T$. 
Consequently, Wan-I2V can use the input image as the first frame and denoise different Gaussian noise samples to spatiotemporally coherent subsequent frames conditioned on it, effectively generating multiple spatiotemporally parallel videos where the person performs different actions. 
This property motivates our method for video CFEs.

\subsection{The BTTF Framework for Video CFEs}

The central task for our BTTF framework is to find the video that best serves as the CFE from the set of spatiotemporally parallel videos generated by Wan-I2V.
To achieve this, we directly optimize the initial latent input $\mathbf{z}_T$ using gradients from the classifier's classification loss.
Our proposed solution consists of two stages (\cref{fig:scheme}), which are detailed below (pseudocode in supplementary material).

\begin{figure*}[t]
  \centering
  \includegraphics[width=1.0\linewidth]{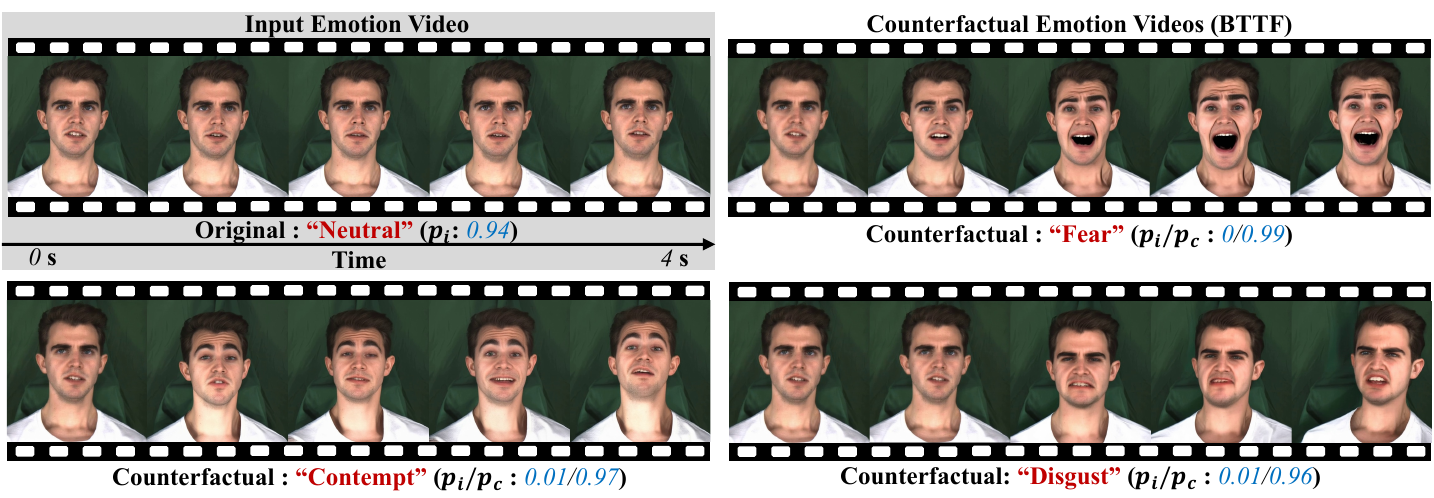}
   \caption{\textbf{CFE videos generated by BTTF for the target emotion classifier E-swin trained on MEAD. } BTTF alters E-swin's prediction on the original input video from ``Neutral'' to target motion classes ``Fear'', ``Contempt''  and ``Disgust'', respectively. The results indicate the strong capacity of BTTF in editing emotion features in a semantically meaningful way.}
   \label{fig:mead_CFE}
\end{figure*}

\textbf{Stage 1: Inversion.}
To ensure that the search remains within the vicinity of the original input video $\mathbf{x}_i$, we first optimize the initial latent $\mathbf{z}_T$ so that it can reconstruct a video similar to $\mathbf{x}_i$. 
To obtain a label for this optimization, we encode $\mathbf{x}_i$ into its VAE latent representation $\mathbf{z}_i = \mathcal{E}(\mathbf{x}_i)$. Then, we iteratively denoise the initial latent $\mathbf{z}_T$ to obtain a noise-free latent $\hat{\mathbf{z}}_0$, which constitutes the prediction for the following reconstruction loss:
\begin{equation}
    \mathcal{L}_I(\hat{\mathbf{z}}_0,\mathbf{z}_i) 
    \;=\; \lVert \hat{\mathbf{z}}_0 - \mathbf{z}_i \rVert_{1}
  \label{eq:loss_inv}
\end{equation}
The reconstruction loss is minimized to optimize $\mathbf{z}_T$, which anchors the second-stage optimization around the neighborhood of $\mathbf{x}_i$. In this stage, the number of denoising steps is fixed to one.

\textbf{Stage 2: CFE generation.}
We further optimize $\mathbf{z}_T$ to induce the target class $y_c$ in the classifier's prediction. 
In this stage, $\hat{\mathbf{z}}_0$ is further decoded by the VAE decoder $\mathcal{D}$ to obtain the video $\hat{\mathbf{x}}_c$, which is then fed into the target classifier to obtain its prediction $\hat{y}$. 
The cross-entropy loss with the target class $y_c$ provides the main driving signal to steer $\mathbf{z}_T$ towards generating a counterfactual video of the target class. 
To keep the generated video $\hat{\mathbf{x}}_c$ in the data manifold of $\mathbf{x}_i$, we introduce a style loss $\mathcal{L}_S$~\cite{gatys2015neural, gatys2015texture}. This loss is the squared Frobenius norm of the difference between the Gram matrices ($G$) of the input and counterfactual videos. That is,  $\mathcal{L}_S(\hat{\mathbf{x}}_c, \mathbf{x}_i) = \frac{1}{N_f\,C^{2}}\sum_{n=1}^{N_f}\lVert G(\hat{\mathbf{x}}_{c,n})-G(\mathbf{x}_{i,n})\rVert_F^2$, where $N_f$ is the number of frames and $C = 3$ because the videos are RGB.
As Gram matrices discard absolute coordinates, this loss term is invariant to in-plane shifts. 
This property makes the style loss compatible with motion feature editing (see~\cref{fig:shape_moving_CFE}). 
The final objective of the CFE generation stage is therefore
\begin{align}
    \mathcal{L}_C(\mathbf{x}_i,\hat{\mathbf{x}}_c,y_{c},\hat{y}) &= -\sum_{k=1}^{K} y_{c,k}\,\log \hat{y}_{k} \notag + \lambda \mathcal{L}_S(\hat{\mathbf{x}}_c, \mathbf{x}_i)
\label{eq:loss_cf}
\end{align}
where $K$ denotes the total number of classes and $\lambda = 1\times10^5$ is a regularization coefficient. 

\textbf{Progressive optimization.}
The considerable depth of the backpropagation chain in our optimization framework leads to an inherent susceptibility to gradient vanishing. 
Two mechanisms mitigate this issue: (i) residual denoising, where, inspired by ResNet blocks~\cite{he2016deep}, we improve gradient flow through denoising step updates of the latent $\mathbf{z}_t$ in residual form, i.e. $\mathbf{z}_{t-1} \approx \mathbf{z}_t - \epsilon_{\phi}(\mathbf{z}_t)$ (up to scheduling coefficients); and (ii) progressive optimization, where we progressively increase the number of denoising steps from $1$ to $N = 15$ during the Stage 2 (CFE generation).

 \section{Experiments and Results}
\label{sec:exp}

\begin{figure*}[t]
  \centering
  \includegraphics[width=1.0\linewidth]{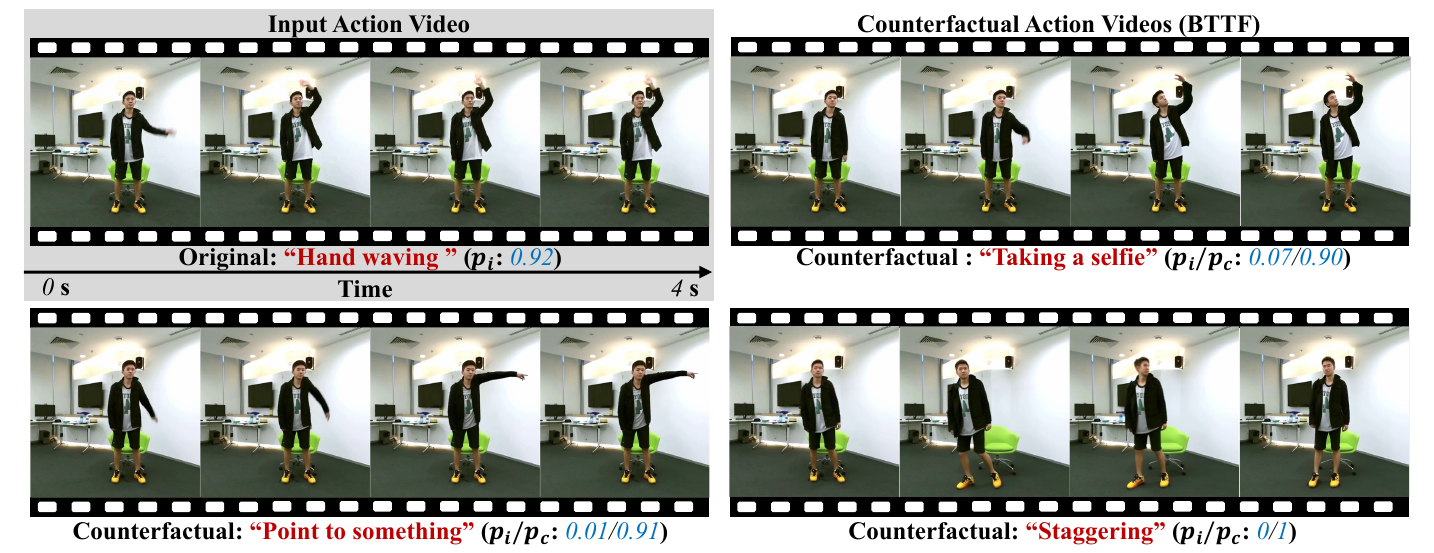}
   \caption{\textbf{Video CFE generated by BTTF for the target video classifier A-swinR trained on NTU RGB+D.} BTTF flips A-swinR's prediction on the original input video from ``Hand waving'' to target action classes ``Taking a selfie'', ``Point to something''  and ``Staggering'', respectively. The results  demonstrate the edits of BTTF for human actions are physically plausible.}
   \label{fig:ntu_CFE}
\end{figure*}

\subsection{Experimental Setup}

\textbf{Datasets}. (i) \textbf{Shape-Moving} (motion classification). This is our synthetic dataset for testing whether video CFE methods are capable of editing spatiotemporal features. Each video contains 65 frames at 16 frames per second (fps) with a spatial resolution of \textbf{224×224}. Inside each video, a single object moves from an arbitrary initial position toward one of the four boundaries (up, down, left, right); the movement direction is used as the class label. Here, we define \textbf{``pure spatiotemporal feature''} as one whose label cannot be inferred from any single frame and requires information from at least two frames with temporal order. Under this definition, movement directions in Shape-Moving are pure spatiotemporal, making the dataset serve as a rigorous benchmark for evaluating the spatiotemporal editing capability of video CFE methods. (ii) \textbf{MEAD}~\cite{wang2020mead} (emotion classification). We uniformly preprocess the original dataset into 33-frame videos at 8 fps, resized to \textbf{512×512}, covering 8 facial emotion classes. (iii) \textbf{NTU RGB+D}~\cite{shahroudy2016ntu}(action classification). We preprocess raw videos to 33 frames at 8 fps, resized to \textbf{448×448}, spanning 60 action classes. 

\textbf{Target classifiers}. We train the state-of-the-art video classifier Video Swin Transformer~\cite{liu2022video} on each dataset. For NTU RGB+D, we train the model using PGD adversarial robust training~\cite{madry2017towards}. The performance evaluation is summarized in~\cref{tab:classifiers_eval}. These classifiers are used as target classifiers to evaluate the quality of CFE methods in subsequent experiments.

\textbf{Diffusion models fine-tuning} We fine-tune\footnote{{\scriptsize Code adapted from \url{https://github.com/modelscope/DiffSynth-Studio/tree/main/examples/wanvideo}}} diffusion models using the LoRA~\cite{hu2022lora} approach.  Specifically, the diffusion models used for explaining classifiers trained on Shape-Moving and NTU RGB+D are fine-tuned directly on their respective training sets to adapt to the domain-specific motion and appearance statistics. For classifiers trained on MEAD, the diffusion model is fine-tuned on the larger face video dataset CelebV-Text~\cite{yu2023celebv}. Since Wan-I2V is originally designed to accept text prompts as one of the conditioning inputs, we employ a class-agnostic fixed prompt template during both fine-tuning and inference to prevent any prompt-related bias in the CFE generation process. For instance, when fine-tuning on Shape-Moving, we use the fixed prompt ``This is a synthetic video'', and the same prompt is applied during CFE generation. This design ensures that the generated CFEs are guided purely by the target classifier’s feedback and the original input video, rather than by linguistic priors embedded in the prompt inputs of diffusion models.

\subsection{Generated CFE Videos}

The CFE videos generated by BTTF for different target video classifiers are shown in~\cref{fig:shape_moving_CFE} ,~\cref{fig:mead_CFE} and~\cref{fig:ntu_CFE}.

\textbf{Spatiotemporal editing capability.} As~\cref{fig:shape_moving_CFE} shows, BTTF generates multiple CFE videos for the motion classifier M-swin by precisely changing the movement directions of the object in the original video, demonstrating that BTTF is capable of manipulating pure spatiotemporal features to produce CFE videos. 

\textbf{Desired properties of video CFE.} Firstly, results for emotion and action classifiers (\cref{fig:mead_CFE} and~\cref{fig:ntu_CFE}) show that generated CFE videos successfully flip the original predictions of target classifiers to their corresponding target classes (\textbf{validity}). Secondly, they preserve original facial identity and scene layout (\textbf{proximity}). Thirdly, edits concentrate on actionable spatiotemporal features, such as facial and body movements (\textbf{actionability}). Fourthly, they are perceptually consistent with original videos (\textbf{realism}). Lastly, generated counterfactual videos are spatiotemporally consistent across frames (\textbf{spatiotemporal consistency}). These results indicate that CFE videos produced by BTTF satisfy the desired properties described in~\cref{sec:intro}. 

\subsection{Ablation Study}

\begin{figure*}[t]
  \centering
  \includegraphics[width=1.\linewidth]{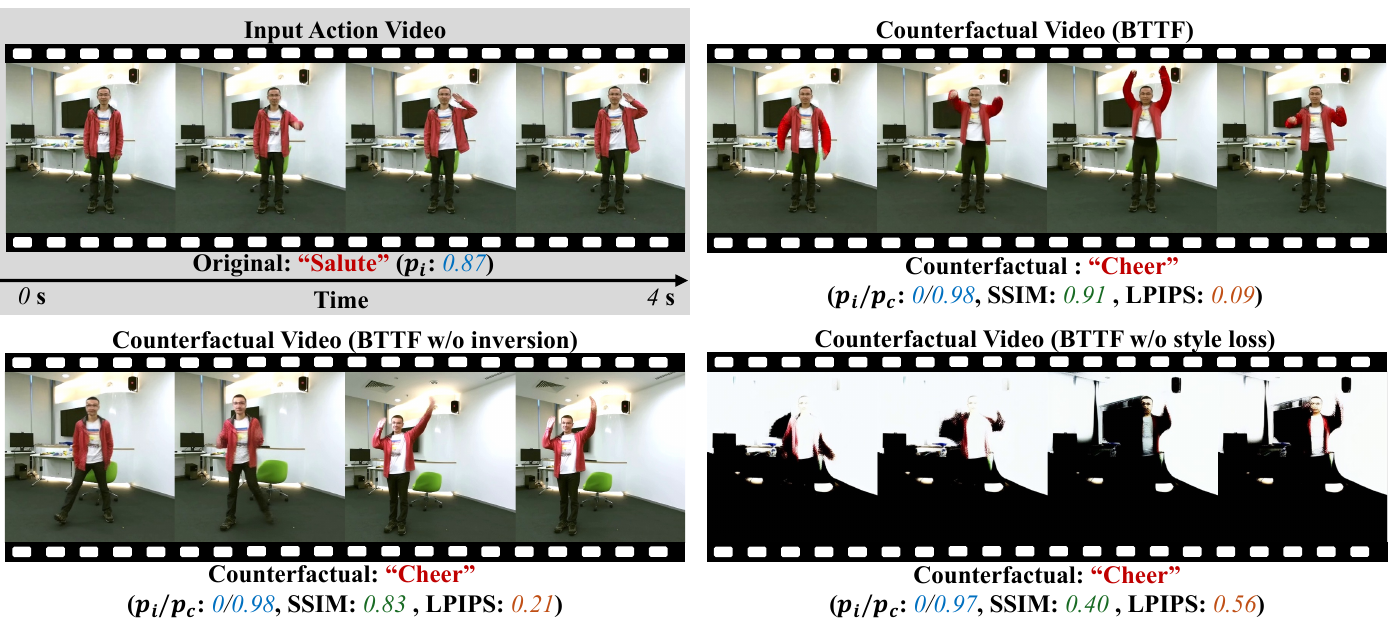}
   \caption{\textbf{Ablation study.} Without inversion, CFE video contains some unnecessary changes (stepping rightward in the bottom left row), validating the effectiveness of inversion in enhancing the proximity of CFE videos. Without style loss, CFE video suffers severe quality degradation (in the bottom right row), validating the effectiveness of style loss regularization in maintaining the realism of CFE videos.}
   \label{fig:ablation}
\end{figure*}

To validate the role of inversion in enhancing the proximity of generated counterfactuals and the role of style regularization in improving visual realism, we conduct an ablation study, as shown in~\cref{fig:ablation}. 

\textbf{Inversion}. When transforming a ``Salute'' video to the target class ``Cheer'', the counterfactual video produced by the complete method preserves the actor's spatial position and directly edits the celebratory action, where both arms are raised in place. 
In contrast, without inversion, the counterfactual introduces an extraneous action: the actor first steps to the right and then raises both arms. 
Although the target classifier assigns the same high confidence (0.98) to both outputs, the complete variant makes fewer edits, indicating that the inversion stage suppresses unnecessary feature changes, producing CFEs that remain closer to the original video.

\textbf{Video style regularization}. Excluding the video style regularization, the generated counterfactual video suffers severe visual quality degradation, demonstrating that the video style regularization is critical for maintaining the perceptual realism of the generated videos during the second-stage optimization.

\subsection{Comparison of Video CFE Methods}

\begin{figure}[t]
  \centering
  \includegraphics[width=\linewidth]{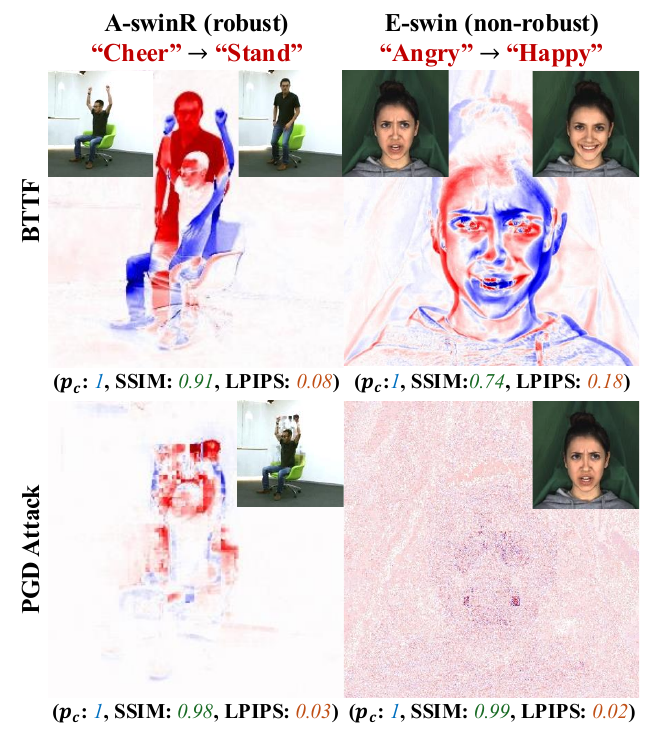}
   \caption{\textbf{Qualitative comparison of methods.} The perutrbations (difference maps) produced by BTTF (upper row) and PGD Attack (bottom row) for A-swinR (left column) and E-swin (right column), respectively. For each column, the left side shows frames extracted from the input video, while the right side shows the corresponding frames at the same timestamps from the generated counterfactual videos.}
   \label{fig:qual_method_com}
\end{figure}

To our knowledge, there are no publicly available methods of video CFEs. Here, we report our comparison results against an adversarial method PGD Attack~\cite{madry2017towards}. Our goal is to provide a valuable reference for the future work of video CFE. The qualitative comparison result of CFE videos for the robust classifier A-swinR is shown in~\cref{fig:qual_method_com}. While PGD Attack achieves lower LPIPS~\cite{zhang2018unreasonable}, higher prediction confidence and SSIM~\cite{wang2004image}, the perturbation it produced is noisy and meaningless, which fails to provide information about model's decision-making logic. In contrast, BTTF depicts a physically plausible action in its CFE video to present the classifier's understanding of the target action. This indicates the limitation of existing metrics in evaluating video CFE work. The quantitative comparison result (\cref{tab:quan_method_com}) of CFE videos for the non-robust classifier E-swin further validates this conclusion. Although our method achieves relatively plausible results, PGD Attack is nearly invincible in these metrics (extremely high flip ratio and SSIM, and correspondingly low LPIPS, FID~\cite{heusel2017gans}, and FVD~\cite{unterthiner2018towards}). This is because the target classifier lacks robustness, meaning that even minor pixel-level tweaks can flip its decision. Therefore, developing a fair and objective quantitative evaluation system remains an open problem in this field.  

\begin{table}
  \caption{\textbf{Quantitative comparison of methods.} Flip rate (FR) evaluates the success rate of generating valid counterfactuals. SSIM, LPIPS and FID are computed between frames of original input video and those of generated counterfactual videos. FVD is computed between original input videos and CFE videos.}
  \label{tab:quan_method_com}
  \centering
  \begin{tabular}{@{}llllll@{}}
    \toprule
    Method  & FR & SSIM & LPIPS & FID & FVD \\
    \midrule
    \multirow{1}{*}{PGD Attack}
      & 1.00    & 0.99    & 0.02    & 0.68    & 0.53 \\
    \addlinespace
    \multirow{1}{*}{BTTF (ours)}
      & 0.99    & 0.81    & 0.17    & 19.03    & 275.44 \\
    \bottomrule
  \end{tabular}
\end{table}

\subsection{Application Case: Finding Spurious Features with BTTF}

\begin{figure}[t]
  \centering
  \includegraphics[width=\linewidth]{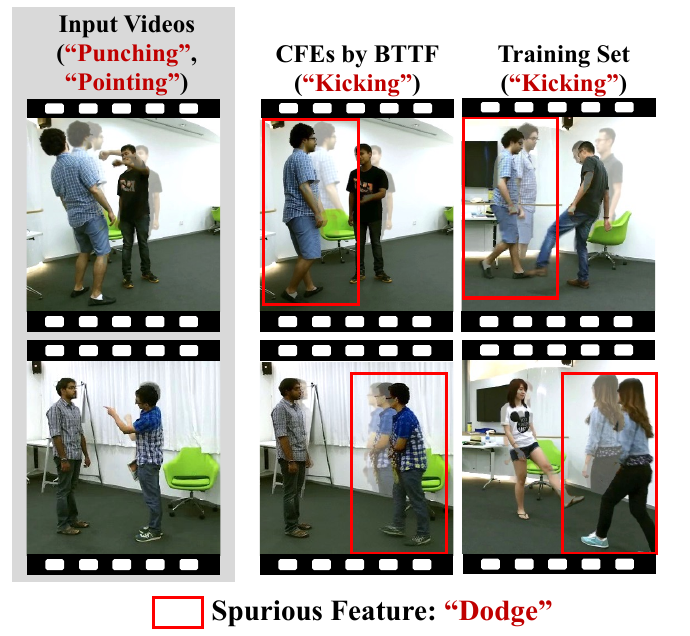}
   \caption{\textbf{Spurious features detection by BTTF.} The leftmost column is original videos, while the middle is generated CFE videos by BTTF with the target class ``kicking'' and the rightmost is ``kicking'' videos in the training set. Inside CFE videos, there are no actual kicking actions but someone moving backward (red boxes). That is because the target classifier A-swinR mistakenly recognizes the dodge movement of the other person as ``kicking'', indicating the great potential of BTTF serving as a debugging tool to detect spurious action features.}
   \label{fig:application_case}
\end{figure}

To further validate the application value of BTTF in detecting failure modes of video classifiers, we employ it to detect spurious features in the target classifier A-SwinR. An example in~\cref{fig:application_case} illustrates a very hidden failure mode of A-SwinR. In the CFE videos produced by BTTF for the target class ``kicking'', we consistently observed no actual kicking actions. Instead, the CFE videos always depict someone stepping backward. A subsequent inspection of the training set reveals the source of this spurious feature. Many ``kicking'' videos depict two-person interactions in which one person performs the kick while the other moves backward to dodge it. The target classifier mistakenly recognizes the retreat action as ``kicking.'' 

 \section{Limitations and Future Work}
\label{sec:lim}

Despite its effectiveness, our framework faces several limitations that offer avenues for future work.
Primarily, video CFEs impose substantial computational demands. The vast parameter space of modern video diffusion models results in high inference latency ($\approx$2h on an 80GB NVIDIA A100 GPU to generate a 4-second counterfactual video). Improving inference efficiency and scaling to longer temporal horizons without compromising generative quality remains a priority. 
Additionally, the current implementation requires domain alignment between the generator and the target classifier via fine-tuning.
Future work will focus on developing multi-domain generators and adaptive mechanisms for cross-domain automation. 
This includes devising mechanisms to automatically tune the hyperparameters of the two-stage generation pipeline to adapt them to new datasets.
Finally, the field lacks standardized evaluation protocols, as conventional metrics like FID/FVD fail to account for semantic or causal validity. 
Establishing new metrics aligned with human perception of spatiotemporal plausibility is critical for advancing explainable video analysis. \section{Conclusion}
\label{sec:conclu}

In this work, we proposed BTTF, an optimization framework that extends CFEs to the video domain.
BTTF is capable of editing actionable spatiotemporal features to alter a target video classifier's prediction, thereby providing human users with insights into a model's internal decision mechanism.
Our experiments demonstrate that the generated counterfactual videos satisfy the key criteria for video CFEs. This ability to manipulate dynamic, semantic features is a critical step towards building more transparent and reliable video-based systems.
Building on this, we demonstrate for the first time that video CFEs can serve as a powerful debugging tool.
Our discovery of a state-of-the-art classifier mistaking a ``dodge'' action for ``kicking'' underscores the practical value of BTTF in model auditing and detecting spurious features.
We believe this work paves the way for more fine-grained analysis of complex spatiotemporal models.

{
    \small
    \bibliographystyle{ieeenat_fullname}
    \bibliography{main}
}

\clearpage
\setcounter{page}{1}
\maketitlesupplementary
\appendix
\renewcommand{\thefigure}{S\arabic{figure}}
\setcounter{figure}{0}
\renewcommand{\thetable}{S\arabic{table}}
\setcounter{table}{0}

\section{Diffusion Model Background}
\label{sec:diff_model_bg}

\subsection{Denoising Diffusion Probabilistic Models}
\label{sec:ddpm}
Denoising diffusion probabilistic models (DDPMs)~\cite{ho2020denoising,song2019generative,song2020improved} consist of two Markov processes: a forward process that gradually adds Gaussian noise to a clean sample $\mathbf{x}_0 \in \mathbb{R}^{n \times c \times h \times w}$ until obtaining a highly-noised sample $\mathbf{x}_T \in \mathbb{R}^{n \times c \times h \times w}$, and a reverse process that removes the noise on $x_T$ to recover the original data distribution. The forward process is defined as:
\begin{equation}
  q(\mathbf{x}_t \mid \mathbf{x}_{t-1}) := \mathcal{N}(\sqrt{\alpha_t}\mathbf{x}_{t-1}, (1 - \alpha_t) I),
  \label{eq:DDPM_forward}
\end{equation}
where $\alpha_t \in \{\alpha_t:1 \leq t \leq T\}$ defines the noise schedule, $I$ is the identity matrix, and $n$, $c$, $h$ and $w$ denote the number, channel, height and width of frames, respectively. Alternatively, given $\mathbf{x}_0$, the noised sample $\mathbf{x}_t$ at any timestep $t$ can be analytically derived from: 
\begin{equation}
  \mathbf{x}_t = \sqrt{\bar{\alpha}_t} \mathbf{x}_0 + \sqrt{1 - \bar{\alpha}_t} \mathbf{\epsilon},
  \label{eq:DDPM_forward_1_step}
\end{equation}
where $\bar{\alpha}_t:= \prod_{i=1}^{t} \alpha_i$ and $\mathbf{\epsilon} \in \mathbb{R}^{n \times c \times h \times w}$ is the Gaussian noise sampled from the standard Gaussian distribution. In the reverse process, a denoiser (e.g., U-Net or Diffusion Transformer) parameterized by $\phi$ is trained to reverse the forward process: 
\begin{equation}
  p_{\phi}(\mathbf{x}_{t-1} \mid \mathbf{x}_t) := \mathcal{N}(\mu_{\phi}(\mathbf{x}_t, t), \Sigma_{\phi}(\mathbf{x}_t,t)),
  \label{eq:DDPM_reverse}
\end{equation}
where $\mu_{\phi}$ and $\Sigma_{\phi}$ are the predicted mean and variance of the data distribution of $\mathbf{x}_t$, respectively. The less-noisy sample $\mathbf{x}_{t-1}$ is drawn from $p_{\phi}(\mathbf{x}_{t-1} \mid \mathbf{x}_t)$ step by step until the noise-free sample $x_0$ is obtained. In most cases, a denoiser estimates the mean $\mu_{\phi}$ by predicting the added noise $\mathbf{\epsilon}_{\phi}$ between $\mathbf{x}_t$ and $\mathbf{x}_0$:
\begin{equation}
  \mu_{\phi}(\mathbf{x}_t, t) = \frac{1}{\sqrt{\alpha_t}} \left( \mathbf{x}_t - \frac{1 - \alpha_t}{\sqrt{1 - \bar{\alpha}_t}} \epsilon_{\phi}(\mathbf{x}_t, t) \right)
  \label{eq:DDPM_reverse_mean}
\end{equation}

\subsection{Denoising Diffusion Implicit Models}
\label{sec:ddim}
In order to speed up the generation process at the inference time, $\mathbf{x}_{t-1}$ can be calculated in a deterministic way using the denoising diffusion implicit model (DDIM)~\cite{song2020denoising} sampler: 
\begin{equation}
  \mathbf{x}_{t-1} = \sqrt{\bar{\alpha}_{t-1}} \hat{\mathbf{x}}_0 + \sqrt{1-\bar{\alpha}_{t-1}} \epsilon_\phi(\mathbf{x}_t, t)
  \label{eq:DDIM}
\end{equation}
where $\hat{\mathbf{x}}_0$ is the clean sample approximated at the timestep $t$ by rewriting \cref{eq:DDPM_forward_1_step}:
\begin{equation}
  \hat{\mathbf{x}}_0 = \frac{\mathbf{x}_t - \sqrt{1-\bar{\alpha}_t} \epsilon_\phi(\mathbf{x}_t, t)}{\sqrt{\bar{\alpha}_t}}
  \label{eq:DDIM_x0}
\end{equation}

\subsection{Classifier-guidance}
\label{sec:CG}
To condition the class of generated samples, Dhariwal et al.~\cite{dhariwal2021diffusion} proposed classifier guidance, where an external classifier $f_\theta$ is introduced into the denoising process of diffusion models. Specifically, the classifier's gradients with respect to noised images $\nabla_{\mathbf{x}_t} \log f_\theta(\mathbf{y}_c \mid \mathbf{x}_t)$ steer the denoising trajectory toward the data manifold of the target class $\mathbf{y}_c \in \mathbb{R}^{k}$ by modifying the predictions of the denoiser: 
\begin{equation}
  \hat{\epsilon}_{\phi}(\mathbf{x}_t, t, \mathbf{y}_c) = \epsilon_{\phi}(\mathbf{x}_t, t) - w \sqrt{1 - \bar{\alpha}_t} {\nabla_{\mathbf{x}_t} \log f_\theta(\mathbf{y}_c \mid \mathbf{x}_t)}
  \label{eq:classifier_guidance}
\end{equation}
where $\hat{\epsilon}_{\phi}(\mathbf{x}_t, t, \mathbf{y}_c)$ is the modified prediction of the denoiser and $w$ is the guidance scale.

\section{Method Details}
\label{sec:method_details}
The optimization framework of BTTF is summarized in~\cref{alg:BTTF_alg}. We define ``denoising loop'' as a flow matching sampling process where the noise-free latent $\hat{\mathbf{z}}_0$ is obtained by recursively running~\cref{eq:DDPM_forward} $n$ times, as summarized in~\cref{alg:denoising_loop}. The whole framework consists of two stages: inversion and CFE generation. In the first stage, the sampled latent $\mathbf{z}_T$ is optimized to be close to the latent representation of the original input video $\mathbf{z}_i$. In the second stage, we progressively optimize the initial latent $\mathbf{z}_T$ to search for a counterfactual video $\hat{\mathbf{x}}_c$ that changes the prediction of the target classifier $f_\theta$ towards the target class $\mathbf{y}_c$. In addition, gradient checkpointing is utilized to mitigate the GPU memory consumption of the proposed method. The hyperparameters of CFE generation experiments for different target classifiers are shown in~\cref{tab:bttf_hyperparams}. Notably, for the target classifier A-swinR, because the semantic differences among action classes in the NTU RGB+D training set may be relatively large (e.g., single-person ``salute'' vs. two-person ``hugging''), the number of inversion iterations, $K_I$, needs to be manually adjusted according to the gap between the class of the original input video and the target class. This empirical heuristic suggests using a smaller $K_I$ when the semantic gap between the original and target classes is large, and a larger $K_I$ otherwise.

\begin{table}[t]
\centering
\caption{\textbf{Hyperparameters used in the BTTF optimization algorithm.}}
\label{tab:bttf_hyperparams}
\resizebox{\columnwidth}{!}{
\begin{tabular}{llll}
\toprule
Hyperparameter & M-swin & E-swin & A-swinR \\
\midrule
\# Inversion iterations $K_I$ & 0 & 40 & 0-40 \\ 
\# CFE generation iterations $K_C$ & 100 & 100 & 100 \\ 
Style loss regularization coefficient $\lambda$ & $1\times10^5$ & $1\times10^5$ & $1\times10^5$ \\ 
Max number of denoising steps $N$ & 15 & 15 & 15 \\ 

\bottomrule
\end{tabular}}
\end{table}

\section{Details of Model Training}
\label{sec:details_model_training}

\subsection{Video Dataset Preprocessing}
\label{sec:video_dataset_pre}

\textbf{Shape-Moving.}
The Shape-Moving dataset is our synthetic video dataset consisting of 64k videos.
Each video contains 65 frames with a spatial resolution of $3 \times 224 \times 224$ (channel $\times$ width $\times$ height) and a frame rate of 16 fps.
The dataset includes four directional categories: ``Up'', ``Down'', ``Left'', and ``Right'', with 16k videos per class.
We randomly split the dataset into training and testing sets using a 4:1 ratio.

\textbf{MEAD.}~\cite{wang2020mead}
The raw MEAD videos have a spatial resolution of $3 \times 1080 \times 1920$. For preprocessing, we first resize the shorter spatial dimension (height) to 512 while preserving the aspect ratio, which is followed by a center crop along the longer dimension to obtain videos with a spatial resolution of $512 \times 512$. In the temporal domain, we perform either downsampling or upsampling so that all videos are normalized to 33 frames, which are then saved at 8 fps, resulting in clips of approximately 4 seconds. The dataset is randomly divided into training and testing sets with a 4:1 split.

\textbf{NTU RGB+D.}~\cite{shahroudy2016ntu}
The preprocessing pipeline for NTU RGB+D follows the same procedure as that used for the MEAD dataset, including spatial resizing and center cropping to $512 \times 512$, as well as temporal normalization to 33 frames at 8 fps.

\subsection{Video Classifiers Training}
\label{sec:details_model_training}

We trained\footnote{\scriptsize Code adapted from \url{https://github.com/open-mmlab/mmaction2}} three target video classifiers, namely M-swin, E-swin, and A-swinR, which are built upon an ImageNet-1K pretrained Video Swin Transformer~\cite{liu2022video} backbone. The training sets, input clip shape ($n \times c \times h \times w$), and number of training epochs, are summarized in~\cref{tab:video_classifiers_training}. 
The input clips are uniformly sampled from videos in the training set using a temporal stride of 2.

\begin{table}[t!]
  \centering
  \caption{\textbf{Training configurations for video classifiers.}}
  \vspace{5pt}
  \label{tab:video_classifiers_training}
  \resizebox{\columnwidth}{!}{
    \begin{tabular}{@{}lcccccc@{}}
        \toprule
        Model & Training Dataset & Input Clip Shape & \# Training Epochs \\
        \midrule
        M-swin  & Shape-Moving & $32 \times 3 \times 224 \times 224$ & 1  \\
        E-swin  & MEAD         & $16 \times 3 \times 512 \times 512$ & 15 \\
        A-swinR & NTU RGB+D    & $16 \times 3 \times 448 \times 448$ & 30 \\
        \bottomrule
    \end{tabular}
    }
\end{table}

\begin{algorithm}
    \caption{BTTF optimization algorithm}
    \label{alg:BTTF_alg}
    \begin{algorithmic}[1]

        \State \textbf{Input:} Original video $\mathbf{x}_i$, target class $y_c$, target video classifier $f_{\theta}$, max number of denoising steps $N$, number of inversion iterations $K_I$, number of CFE generation iterations $K_C$, the objective functions $\mathcal{L}_I$ and $\mathcal{L}_C$
        \State \textbf{Output:} CFE video $\mathbf{x}_c$

        \Statex
        \State $\mathbf{I} \gets \text{first\_frame}(\mathbf{x}_i)$ \Comment{\textcolor{blue!60!black}{Extract the original first frame}}
        \State $\mathbf{I}_C \gets \text{CLIP}(\mathbf{I})$ \Comment{\textcolor{blue!60!black}{Extract CLIP embedding}}
        \State $\mathbf{I}_p \gets \text{zero\_pad\_to\_video}(\mathbf{I})$ \Comment{\textcolor{blue!60!black}{Zero-pad the first frame}}
        \State $\mathbf{I}' \gets \mathcal{E}(\mathbf{I}_p)$ \Comment{\textcolor{blue!60!black}{Encode the zero-padded first frame}}
        \State $\mathbf{z}_i \gets \mathcal{E}(\mathbf{x}_i)$ \Comment{\textcolor{blue!60!black}{Encode the original video}}
        \State $\mathbf{z}_T \gets \text{sample\_normal\_like}(\mathbf{z}_i)$ \Comment{\textcolor{blue!60!black}{Sample the latent}}
        \State $\text{optim} \gets \text{AdamW}(\mathbf{z}_T)$ \Comment{\textcolor{blue!60!black}{Initialize the optimizer}}
        
        \Statex
        \Comment{\textcolor{orange}{\textbf{Stage 1: Inversion}}}
        \State $n \gets 1$
        \For{$k \gets 1, \dots, K_I$}
            \State $\text{optim}.\text{zero\_grad}()$
            \State $\hat{\mathbf{z}}_0 \gets \text{denoising\_loop}(\mathbf{z}_T, \mathbf{I}', \mathbf{I}_C, n)$
            \State $l \gets \mathcal{L}_I(\hat{\mathbf{z}}_0, \mathbf{z}_i)$
            \State $l.\text{backward}()$
            \State $\text{optim}.\text{step}()$ \Comment{\textcolor{blue!60!black}{Update $\mathbf{z}_T$}}
        \EndFor

        \Statex
        \Comment{\textcolor{darkgreen}{\textbf{Stage 2: CFE generation}}}
        \For{$n \gets 1, \dots, N$}
            \For{$k \gets 1, \dots, K_C$}
                \State $\text{optim}.\text{zero\_grad}()$
                \State $\hat{\mathbf{z}}_0 \gets \text{denoising\_loop}(\mathbf{z}_T, \mathbf{I}', \mathbf{I}_C, n)$
                \State $\hat{\mathbf{x}}_c \gets \mathcal{D}(\hat{\mathbf{z}}_0)$
                \State $\hat{y} \gets f_{\theta}(\hat{\mathbf{x}}_c)$
                \State $l \gets \mathcal{L}_C(\mathbf{x}_i,\hat{\mathbf{x}}_c,y_{c},\hat{y})$
                \State $l.\text{backward}()$
                \State $\text{optim}.\text{step}()$ \Comment{\textcolor{blue!60!black}{Update $\mathbf{z}_T$}}
            \EndFor
        \EndFor
        
        \Statex
        \State $\mathbf{x}_c \gets \mathcal{D}(\text{denoising\_loop}(\mathbf{z}_T, \mathbf{I}', \mathbf{I}_C, N))$
        \State \textbf{return} $\mathbf{x}_c$

    \end{algorithmic}
\end{algorithm}

\begin{algorithm}
    \caption{Denoising loop used in Algorithm~\ref{alg:BTTF_alg}}
    \label{alg:denoising_loop}
    \begin{algorithmic}[1]
        \State \textbf{Input:} Initial latent $\mathbf{z}_T$, encoded first frame $\mathbf{I}'$, CLIP embedding $\mathbf{I}_C$, number of denoising steps $n$
        \State \textbf{Output:} Denoised latent $\hat{\mathbf{z}}_0$
        \Statex

        \State $\mathbf{z} \gets \mathbf{z}_T$
        \For{$t \gets n, \dots, 1$}
            \State $\boldsymbol{\epsilon} \gets \boldsymbol{\epsilon}_{\phi}(\mathbf{z}, t, \mathbf{I}', \mathbf{I}_C)$
            \State $\mathbf{z} \gets 
            \sqrt{\bar{\alpha}_{t-1}}
            \dfrac{\mathbf{z} - \sqrt{1-\bar{\alpha}_t}\,\boldsymbol{\epsilon}}{\sqrt{\bar{\alpha}_t}}
            +
            \sqrt{1-\bar{\alpha}_{t-1}}\,\boldsymbol{\epsilon}$
        \EndFor
        \State $\hat{\mathbf{z}}_0 \gets \mathbf{z}$
        \State \textbf{return} $\hat{\mathbf{z}}_0$
    \end{algorithmic}
\end{algorithm}

\section{Evaluation Details}
\label{sec:eval_details}

\subsection{Metrics}
\label{sec:metrics}

\textbf{Flip Rate (FR)} measures the effectiveness of counterfactual examples in changing the classifier's prediction to the target class. Given a counterfactual video $x_c$, a target classifier $f_\theta$, and a target class $y_c$, FR is defined as:
\begin{equation}
    \mathrm{FR}(\mathbf{x}_c, \mathbf{y}_c) = \frac{1}{M} \sum_{m=1}^{M} \mathbb{I}\big[f_\theta(\mathbf{x}_c) = \mathbf{y}_c],
\end{equation}
where $\mathbb{I}$ denotes the indicator function and $M$ is the total number of samples. A higher FR indicates more effective counterfactual interventions.

\textbf{SSIM}~\cite{wang2004image} quantifies perceptual similarity between two corresponding frames in input and counterfactual videos based on luminance, contrast, and structural information. For two frames $\mathbf{x}_{i,n}$ and $\mathbf{x}_{c,n}$, SSIM is computed as:
\begin{equation}
    \mathrm{SSIM}(\mathbf{x}_{i,n}, \mathbf{x}_{c,n}) = 
    \frac{(2\mu_{i,n} \mu_{c,n} + C_1)(2\sigma_{ic,n} + C_2)}
    {(\mu_{i,n}^2 + \mu_{c,n}^2 + C_1)(\sigma_{i,n}^2 + \sigma_{c,n}^2 + C_2)},
\end{equation}
where $\mu_{i,n}$, $\mu_{c,n}$ are means, $\sigma_{i,n}^2$, $\sigma_{c,n}^2$ are variances, $\sigma_{ic,n}$ is the covariance, and the constants $C_1$ and $C_2$ are stabilizing terms introduced to avoid numerical instability when the denominators are close to zero. Here, SSIM is averaged across frames of input and counterfactual videos: $\mathrm{SSIM}(\mathbf{x}_i,\mathbf{x}_c)=\frac{1}{N_f}\sum_{n=1}^{N_f}\mathrm{SSIM}(\mathbf{x}_{i,n}, \mathbf{x}_{c,n})$, where $N_f$ is the number of video frames. Higher values indicate higher perceptual similarity. 

\textbf{LPIPS}~\cite{zhang2018unreasonable} measures perceptual similarity using deep network features. Here, we use AlexNet~\cite{krizhevsky2012imagenet}. Given two images, LPIPS is defined as:
\begin{equation}
    \mathrm{LPIPS}(\mathbf{x}_{i,n}, \mathbf{x}_{c,n}) = 
    \sum_{l} w_l \, \left\| \tilde{f}_{A,l}(\mathbf{x}_{i,n}) - \tilde{f}_{A,l}(\mathbf{x}_{c,n}) \right\|_2^2,
\end{equation}
where $f_{A,l}$ denotes features extracted at layer $l$ of AlexNet, $\tilde{f}_{A,l}$ is the normalized feature vector, and $w_l$ are learned weights. Here, LPIPS is averaged across frames of input and counterfactual videos: $\mathrm{LPIPS}(\mathbf{x}_i,\mathbf{x}_c)=\frac{1}{N_f}\sum_{n=1}^{N_f}\mathrm{LPIPS}(\mathbf{x}_{i,n}, \mathbf{x}_{c,n})$. Lower values indicate higher perceptual similarity. 

\textbf{FID}~\cite{heusel2017gans} measures the discrepancy between real and generated images by comparing their Inception-V3 feature distributions. Let $f_I(\mathbf{x})$ denote the feature vector extracted from the Inception-V3 network~\cite{szegedy2016rethinking}. We assume that the feature vectors of real images and generated images follow Gaussian distributions $\mathcal{N}(\mu_r,\Sigma_r)$ and $\mathcal{N}(\mu_g,\Sigma_g)$, respectively. Under this assumption, FID corresponds to the squared 2-Wasserstein distance between these two Gaussian distributions, given by
\begin{equation}
\label{eq:fid}
\begin{split}
W_2^2\!\big(\mathcal{N}(\mu_r,\Sigma_r), \mathcal{N}(\mu_g,\Sigma_g)\big)
= \|\mu_r - \mu_g\|_2^2 \\
\quad + \operatorname{Tr}\!\left(
    \Sigma_r + \Sigma_g - 2(\Sigma_r \Sigma_g)^{1/2}
  \right),
\end{split}
\end{equation}
where $\operatorname{Tr}$ denotes the trace operator (the sum of the diagonal elements of a matrix).
Lower values indicate that the generated images more closely match the real data distribution.

\textbf{FVD}~\cite{unterthiner2018towards} extends FID from images to videos by applying~\cref{eq:fid} to spatiotemporal features extracted with an I3D network~\cite{carreira2017quo}. As in FID, the real and generated video features are modeled as Gaussian distributions, and FVD evaluates the discrepancy between them. Lower values indicate more realistic and temporally coherent video generation.

\subsection{Comparison with CG-based methods}
\label{sec:com_with_CG}

To demonstrate the limitations of applying image-based CFE methods to videos, we adapt the classifier guidance (CG) strategy, commonly used in image-based CFE methods~\cite{jeanneret2022diffusion,augustin2022diffusion,weng2024fast,sobieski2024rethinking},to the video setting and compare it with our proposed BTTF. We construct three CG baselines that progressively incorporate more temporal information: CG-Frame, which edits each frame independently; CG-Video-Mid, which applies video-classifier gradients but starts denoising from a half-noisy state; and CG-Video, which performs full-noise initialization and video-level guidance.

As shown in~\cref{tab:quan_method_com2}, all CG variants struggle to generate high-quality video counterfactuals. CG-Frame fails to maintain temporal consistency, resulting in extremely poor FVD despite reasonable frame-level SSIM. CG-Video-Mid partially improves temporal coherence by using video-level gradients, but the partial-noise initialization limits its ability to introduce coherent spatiotemporal edits. CG-Video, the strongest CG baseline, shows better temporal alignment, yet its FID and LPIPS remain far worse than ours, indicating that simply extending image-based classifier-guidance to videos cannot reliably edit spatiotemporal features.

In contrast, BTTF clearly outperforms CG variants across most metrics, achieving by far the lowest FID (27.49) and FVD (250.11). These results show that BTTF generates counterfactual videos that are simultaneously realistic, temporally stable, and semantically meaningful. 

Overall, this comparison confirms that video counterfactual explanations cannot be obtained by naively adapting image-based classifier-guidance methods, and that our BTTF framework is specifically effective at producing coherent, high-fidelity spatiotemporal counterfactuals.
\begin{table}
  \caption{\textbf{Quantitative comparison of methods.} BTTF is evaluated against three adapted versions of the image-based classifier-guidance (CG) strategy to validate the limitations of image-based CFE methods to video. CG-Frame applies classifier-guidance independently to each video frame; CG-Video-Mid uses video-classifier gradients but begins denoising from a half-noisy state; and CG-Video performs full-noise initialization with video-level guidance. The benchmark dataset and the target classifier are Shape-Moving and M-swin, respectively.}
  \label{tab:quan_method_com2}
  \centering
  \resizebox{\columnwidth}{!}{
      \begin{tabular}{@{}llllll@{}}
        \toprule
        Method  & FR ($\uparrow$) & SSIM ($\uparrow$) & LPIPS ($\downarrow$) & FID ($\downarrow$) & FVD ($\downarrow$) \\
        \midrule
        \multirow{1}{*}{CG-Frame}
          & 0.28    & 0.90    & 0.28    & 160.48    & 2072.14 \\
        \addlinespace
        \multirow{1}{*}{CG-Video-Mid}
          & 1.00    & 0.95    & \textbf{0.14}    & 145.22    & 1796.21 \\
        \addlinespace
        \multirow{1}{*}{CG-Video}
          & \textbf{1.00}    & 0.94    & 0.24    & 68.01    & 682.23 \\
        \addlinespace
        \multirow{1}{*}{BTTF (ours)}
          & 0.97    & \textbf{0.95}    & 0.20    & \textbf{27.49}    & \textbf{250.11} \\
        \bottomrule
      \end{tabular}
  }
\end{table}

\section{Qualitative results }
\label{sec:failure_cases}

\begin{figure*}[t]
  \centering
  \includegraphics[width=1.0\linewidth]{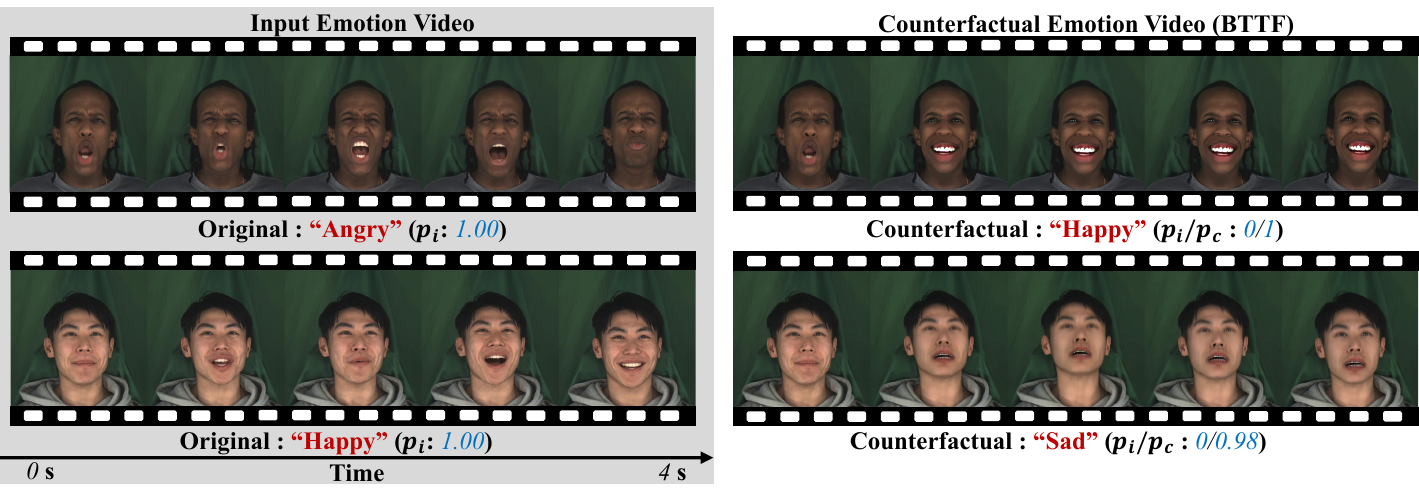}
   \caption{\textbf{Video CFEs generated by BTTF for the target video classifier E-swin trained on MEAD.}}
   \label{fig:emo_CFE}
\end{figure*}

\begin{figure*}[t]
  \centering
  \includegraphics[width=1.0\linewidth]{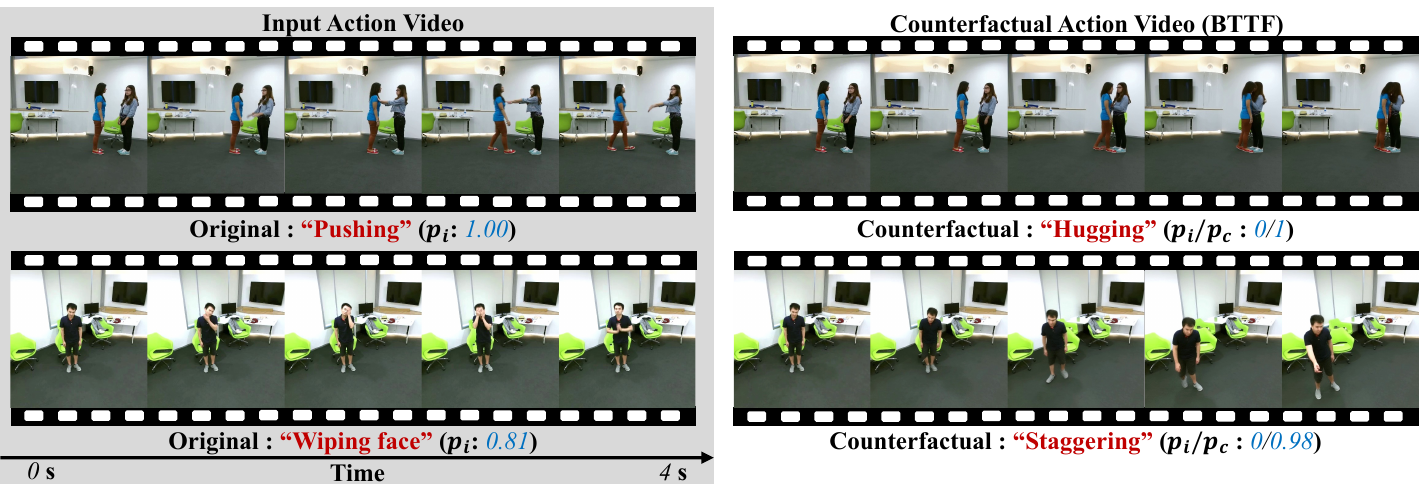}
   \caption{\textbf{Video CFEs generated by BTTF for the target video classifier A-swinR trained on NTU RGB+D.}}
   \label{fig:act_CFE}
\end{figure*}

More generated video CFEs by BTTF are shown in~\cref{fig:emo_CFE} and~\cref{fig:act_CFE}.

\FloatBarrier

\end{document}